\documentclass[lettersize,journal]{IEEEtran}

\usepackage{amsmath,amsfonts}
\usepackage{algorithmic}
\usepackage{algorithm}
\usepackage{array}
\usepackage[caption=false,font=normalsize,labelfont=sf,textfont=sf]{subfig}
\usepackage{textcomp}
\usepackage{stfloats}
\usepackage{url}
\usepackage{verbatim}
\usepackage{graphicx}
\usepackage{cite}
\usepackage{multirow}
\usepackage{booktabs}
\usepackage[pagebackref=true,breaklinks=true,letterpaper=true,colorlinks,bookmarks=false, citecolor=blue]{hyperref}
\usepackage{tabularx}
\usepackage{adjustbox}
\ifCLASSOPTIONcompsoc
  \usepackage[nocompress]{cite}
\else
  \usepackage{cite}
\fi

\ifCLASSINFOpdf
\else
\fi

\begin{document}

\title{FSCA-Net: Feature-Separated Cross-Attention Network for Robust Multi-Dataset Training}

\author{Yuehai Chen, 
         Jing Yang,
         Jian Lan,
         Yi Si,
         Shaoyi~Du,
         Badong Chen%
\IEEEcompsocitemizethanks{\IEEEcompsocthanksitem 
Yuehai~Chen, Jing Yang, and Yi Shi are with School of Automation Science and Engineering, Xi’an Jiaotong University, Xi’an 710049, China. E-mail: cyh0518@stu.xjtu.edu.cn, jasmine1976@xjtu.edu.cn and shiyi@mail.xjtu.edu.cn.
\IEEEcompsocthanksitem  Jian Lan  are with the Institute for Low-Altitude Regulation,
School of Electronics and Information Engineering, Xi'an Jiaotong University, Xi'an 710049, China (e-mail: lanjian@mail.xjtu.edu.cn).
\IEEEcompsocthanksitem Badong Chen and Shaoyi~Du are with Institute of Artificial Intelligence and Robotics, Xi’an Jiaotong University, Xi’an, Shanxi 710049, China.\\
E-mail: chenbd@mail.xjtu.edu.cn, dushaoyi@gmail.com
\IEEEcompsocthanksitem Corresponding author: Jing Yang and Jian Lan. E-mail: jasmine1976@xjtu.edu.cn and lanjian@mail.xjtu.edu.cn.
}}

\maketitle

\begin{abstract}
Crowd counting plays a vital role in public safety, traffic regulation, and smart city management. However, despite the impressive progress achieved by CNN- and Transformer-based models, their performance often deteriorates when applied across diverse environments due to severe domain discrepancies. Direct joint training on multiple datasets, which intuitively should enhance generalization, instead results in negative transfer, as shared and domain-specific representations become entangled. To address this challenge, we propose the Feature Separation and Cross-Attention Network (FSCA-Net), a unified framework that explicitly disentangles feature representations into domain-invariant and domain-specific components. A novel cross-attention fusion module adaptively models interactions between these components, ensuring effective knowledge transfer while preserving dataset-specific discriminability. Furthermore, a mutual information optimization objective is introduced to maximize consistency among domain-invariant features and minimize redundancy among domain-specific ones, promoting complementary shared–private representations. Extensive experiments on multiple crowd counting benchmarks demonstrate that FSCA-Net effectively mitigates negative transfer and achieves state-of-the-art cross-dataset generalization, providing a robust and scalable solution for real-world crowd analysis.
\end{abstract}

\begin{IEEEkeywords}
Crowd Counting, Cross-Domain Generalization, Feature Separation, Cross-Attention, Mutual Information Regularization.
\end{IEEEkeywords}

\markboth{Journal of \LaTeX\ Class Files,~Vol.~14, No.~8, August~2015}%
{Chen \MakeLowercase{\textit{et al.}}: FDGC-DG: Flaw Detection-Based Generative Contrastive Domain Generalization for Zero-Shot Face Anti-Spoofing}

\IEEEdisplaynontitleabstractindextext

%
\IEEEpeerreviewmaketitle

\ifCLASSOPTIONcompsoc
\IEEEraisesectionheading{\section{Introduction}\label{sec:introduction}}
\else
\section{Introduction}
\label{sec:introduction}
\fi

\IEEEPARstart{C}{R}owd counting, which aims to estimate the number and spatial distribution of people in images, has become a fundamental task in computer vision with broad applications in intelligent video surveillance, urban planning, and public safety management. Accurate crowd counting not only enables real-time monitoring of large-scale public gatherings but also provides crucial support for traffic regulation, emergency evacuation, and smart city construction \cite{TNN1, TNNLS2}. With the rapid advancement of deep learning, convolutional neural networks (CNNs) and vision transformers (ViTs) have achieved remarkable success in extracting discriminative features for crowd density estimation, significantly improving the state of the art on individual datasets \cite{TIP9, TIP10}.

Despite these achievements, crowd counting in real-world scenarios often requires handling diverse environments beyond a single dataset \cite{TNNLS3}. This motivates the need for joint training across multiple datasets, which is appealing for two main reasons. First, leveraging a larger and more diverse collection of training samples is expected to improve generalization ability, aligning better with practical deployment where crowd scenes vary drastically in density, perspective, and background. Second, constructing a sufficiently large, high-quality dataset from scratch is prohibitively expensive, since density map annotations demand labor-intensive and time-consuming manual labeling. Thus, combining existing datasets provides a pragmatic and cost-effective pathway toward scalable and robust crowd counting models.

\begin{table}[t]
    \centering
    \caption{Performance comparison using BL\cite{BL} baseline between single-dataset training and joint training on datasets A and B. Metrics are reported in MAE/MSE.}
    \begin{tabular}{c|ccc}
        \toprule
        Training Setting & A (MAE/MSE) & B (MAE/MSE) \\
        \midrule
        A only & 62.8/101.8 & 15.9/25.8 \\
        B only & 138.1/228.1 & 7.7/12.7 \\
        A+B    & 65.3/107.9 & 10.9/16.8 \\
        \bottomrule
    \end{tabular}
    \label{tab:single_vs_joint}
\end{table}

However, a major challenge arises when models are jointly trained on heterogeneous datasets or deployed across unseen domains. Although CNN- and ViT-based models achieve excellent performance within individual datasets, their generalization ability deteriorates severely in cross-dataset scenarios due to substantial distributional gaps \cite{TNNLS4, TNNLS5}. As shown in Table~\ref{tab:single_vs_joint}, joint training on multiple datasets not only fails to yield performance gains but also leads to negative transfer, where accuracy on both source and target domains is degraded. This observation highlights the inherent conflicts among datasets and raises a fundamental question: how can we design models that extract universally transferable features while avoiding contamination from dataset-specific biases?

\begin{figure}[t]
    \begin{center}
    \includegraphics[width=1.0\linewidth]{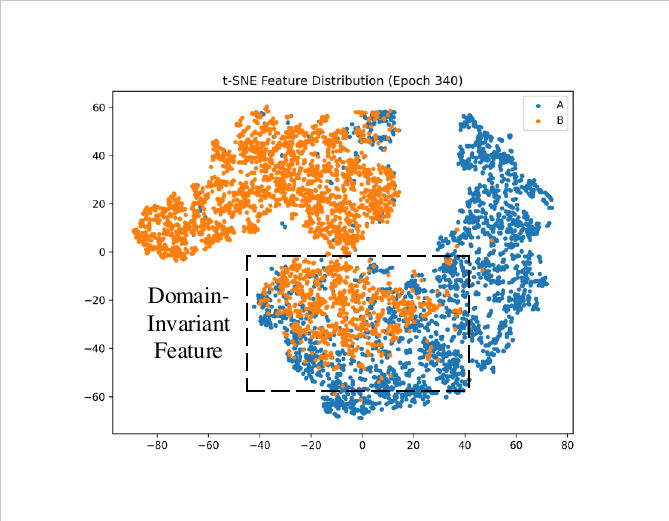}
    \end{center}
   \caption{t-SNE visualization of features extracted from multiple datasets using a baseline model. The features from different datasets are highly entangled, indicating that domain-invariant crowd representations (e.g., density patterns, head shapes) are mixed with domain-specific artifacts (e.g., background textures, illumination, camera perspectives). This entanglement leads to poor cross-dataset generalization, motivating our proposed feature separation strategy.}
    \label{fig:intri}
    \end{figure}

A closer investigation reveals that one of the fundamental reasons behind this poor cross-domain performance lies in the entanglement of dataset-specific characteristics within the learned feature space. In conventional crowd counting networks, low-level and high-level features are extracted jointly, inevitably coupling domain-invariant crowd representations (e.g., density patterns, head shapes) with domain-specific artifacts (e.g., background textures, illumination conditions, camera resolutions). As illustrated in the t-SNE visualization of Fig.~\ref{fig:intri}, features from different datasets overlap in a highly entangled manner, indicating that dataset-invariant and dataset-specific information cannot be effectively disentangled. When the model is deployed in a new domain, these domain-specific cues become unreliable or misleading, ultimately causing significant performance degradation.

To mitigate this problem, several strategies have been explored. Adversarial domain alignment methods attempt to bridge dataset gaps by learning indistinguishable feature distributions across domains, but such approaches often overlook the nuanced balance between shared and private knowledge, leading to over-alignment and loss of domain-specific cues. Meta-learning–based adaptation enhances rapid generalization to unseen domains, yet its reliance on episodic training and limited adaptation capacity constrains effectiveness in complex crowd scenes. Consequently, existing approaches remain insufficient to fully resolve the conflict between domain robustness and dataset specialization.

In this paper, we propose the Feature-Separated Cross-Attention Crowd Counting Network (FSCA-Net), a unified framework designed for multi-dataset crowd counting. Our method explicitly disentangles the feature space into domain-invariant and domain-specific components, enabling the model to capture transferable crowd patterns while isolating dataset-specific variations. A novel cross-attention fusion module facilitates effective interaction between these components, ensuring adaptability without sacrificing discriminability. Furthermore, mutual information optimization is employed to maximize consistency among domain-invariant features while minimizing redundancy among domain-specific ones, thereby reinforcing complementary roles of shared and private representations. Together, these designs allow FSCA-Net to achieve superior cross-dataset generalization while maintaining competitive in-domain accuracy. The key contributions of this work can be summarized as follows:  

\begin{itemize}
    \item \textbf{Disentangled domain representation learning}: We propose a feature separation strategy that explicitly decomposes the backbone representation into domain-invariant and domain-specific subspaces. Guided by orthogonality and adversarial constraints, this design reduces redundant encoding and isolates dataset-specific artifacts, thereby improving the robustness and transferability of learned features across diverse domains.  
    \item \textbf{Cross-attention–based adaptive fusion}: To balance generalization and specialization, a novel cross-attention fusion module is introduced. By treating domain-invariant features as queries and domain-specific features as keys and values, the module dynamically captures conditional dependencies between shared and private representations, achieving adaptive and interpretable cross-domain interaction.  
    \item \textbf{Information-theoretic regularization for complementary learning:} We design a mutual information optimization objective that maximizes the shared information among domain-invariant features while minimizing redundancy among domain-specific ones. This encourages the learned subspaces to be complementary and non-overlapping, ensuring consistent generalization performance in both seen and unseen domains.  
\end{itemize}

\section{Related Work}

\subsection{Crowd Counting}

Crowd counting is commonly performed by employing neural networks to generate a density map, which can be integrated to obtain the total count of individuals within an image \cite{TIP8, TIP9, TIP10}. To enhance prediction accuracy, a considerable body of research has focused on developing sophisticated network architectures that leverage multi-scale and contextual feature information. For instance, MCNN employs three independent branches to estimate crowd counts from different perspectives \cite{MCNN}, while Switching-CNN extends this idea by introducing a switch classifier to dynamically select the most suitable branch for density estimation \cite{switch-CNN}. More recent models, such as M-SFANet \cite{M-SFANet} and SASNet \cite{SASNet}, have proposed novel mechanisms for encoding multi-scale contextual information and mitigating scale variation. MAN \cite{MAN} integrates global attention from a vanilla transformer, learnable local attention, and instance-level attention into the counting framework, leading to improved performance. Similarly, FFDBC \cite{FFDBC} repurposes point annotations to further enhance model accuracy. While these approaches demonstrate the effectiveness of advanced network designs in improving counting performance, they generally assume that the training and testing data are drawn from the same distribution. This assumption often fails in real-world applications, where significant domain shifts can cause substantial performance degradation.

In parallel, a complementary line of work seeks to bypass the explicit construction of density maps by exploiting point annotations more directly. For example, BL \cite{BL} introduces a point-wise loss between ground-truth annotations and aggregated dot predictions derived from the estimated density map. NoiseCC \cite{NoiseCC} explicitly models annotation noise as a Gaussian random variable and incorporates it into the loss function through a probability density approximation. DM-Count \cite{DM-Count} formulates counting as a balanced optimal transport problem with an $L_2$ cost, while its extensions, including GL \cite{GL} and UOT \cite{UOT}, relax this assumption using unbalanced optimal transport. ChfL \cite{ChfL}, in contrast, transforms density maps into the frequency domain and leverages the characteristic function for improved representation, yielding superior results. Although these loss-based approaches have achieved remarkable accuracy, they also rely on the assumption of distributional consistency between training and testing data. Consequently, their performance deteriorates significantly when applied to unseen domains. This highlights the importance of cross-domain evaluation as a critical step toward developing crowd counting models that are robust and generalizable in real-world scenarios.

\subsection{Domain Adaptation} 
Domain adaptation has emerged as a widely used strategy to address the domain shift problem in crowd counting. Domain shift arises when the distribution of the training data significantly differs from that of the test data, leading to degraded model performance. To alleviate this issue, various domain adaptation approaches have been developed. For instance, \cite{synthetic} employ a GAN-based framework to adapt models trained on synthetic datasets to real-world images. However, this method requires prior knowledge of the crowd density distribution in the target domain to carefully match synthetic training samples, in order to avoid negative adaptation effects \cite{AdaCrowd}. Similarly, \cite{Neuron} introduce neuron transformation techniques to explicitly model domain shifts.

Other approaches adopt alternative strategies. For example, \cite{cmot} utilize a momentum-based updating mechanism to learn cross-domain mappings, achieving state-of-the-art performance in zero-shot cross-domain crowd counting. Meanwhile, \cite{AdaCrowd} propose adapting the model by leveraging one or more unlabeled images from the target domain. Although these methods demonstrate promising results, they generally rely on access to target-domain data—either labeled or unlabeled—which is often costly and impractical in real-world crowd counting applications where target scenes are highly dynamic and continuously changing.

\subsection{Domain Generalization} 

Compared to domain adaptation, domain generalization aims to develop models that can generalize well across multiple unseen domains without requiring access to target-domain data. This property makes it particularly suitable for crowd counting, where test data distributions often vary significantly across scenes. Existing domain generalization methods can be broadly categorized into two groups: data diversification and domain alignment.

Data diversification methods focus on increasing the variability of the source domains to improve model robustness. For example, \cite{gan} introduce an adversarial training strategy to augment training data, while \cite{cgan} employ a conditional generative adversarial network to synthesize samples from pseudo-novel domains. On the other hand, domain alignment methods aim to minimize the discrepancy among source-domain features, thereby learning domain-invariant representations. For instance, \cite{domain169} and \cite{yang2017pairwise} propose aligning feature distributions across domains, with the latter leveraging a deep cross-modal hashing framework to generate compact hash codes that effectively capture intrinsic relationships across modalities.

While these approaches demonstrate promising results, they typically focus on either domain adaptation or domain generalization in isolation. In contrast, we propose a novel framework for cross-scene crowd counting that unifies the advantages of both paradigms. To the best of our knowledge, this is the first attempt to design a unified method that jointly leverages domain generalization and domain adaptation to address the challenges of cross-scene crowd counting, thereby improving robustness and adaptability in real-world applications.

\subsection{Mutual Information in Deep Learning}

Mutual information (MI) is a fundamental concept in information theory that quantifies the statistical dependency between two random variables. \cite{MI2016} first introduced MI-related optimization into deep learning, after which a large body of research demonstrated that maximizing MI can effectively enhance feature representation learning \cite{MI1, MI2, MI3}. However, directly estimating MI in high-dimensional feature spaces is computationally intractable. To address this challenge, many studies have developed variational bounds to approximate the true MI \cite{MI4, CLUB}.

In this work, we leverage MI-based regularization to guide feature separation and domain robustness. Specifically, we design two complementary MI objectives: a lower bound and an upper bound. The MI lower bound is used to encourage stronger correlation among domain-invariant features across different domains, thereby enhancing feature consistency and improving adaptation capability. In contrast, the MI upper bound acts as a penalty on domain-specific features, reducing redundancy and limiting their correlation across domains, which promotes better generalization by preventing spurious alignment. By combining these two constraints, the proposed framework simultaneously enhances domain-invariant representation learning and suppresses domain-specific entanglement. Moreover, our design provides a natural bridge between domain generalization and domain adaptation: if unlabeled target-domain samples are available, the generalization strategy can be seamlessly converted into an adaptation method by replacing the MI upper bound with the MI lower bound to explicitly align features with the target distribution.

\section{Proposed Method}
\label{proposed method}

To tackle the challenges of occlusion, scale variation, and distributional shifts across datasets, we design a feature separation framework as the foundation of our proposed method. Specifically, the framework explicitly disentangles representations into two complementary components: domain-invariant (DI) features and domain-specific (DS) features. The DI features capture semantic and structural information that can be generalized across different datasets, while the DS features encode dataset-specific characteristics such as background textures, illumination conditions, and perspective distortions. By jointly leveraging these two components, the framework preserves transferable knowledge across datasets while selectively retaining distinctive domain properties. This design effectively mitigates distributional conflicts, improves the adaptability of learned representations, and enhances the robustness of crowd counting under cross-domain scenarios.

\subsection{Overall Framework}

\begin{figure*}[t]
\begin{center}
\includegraphics[width=1.0\linewidth]{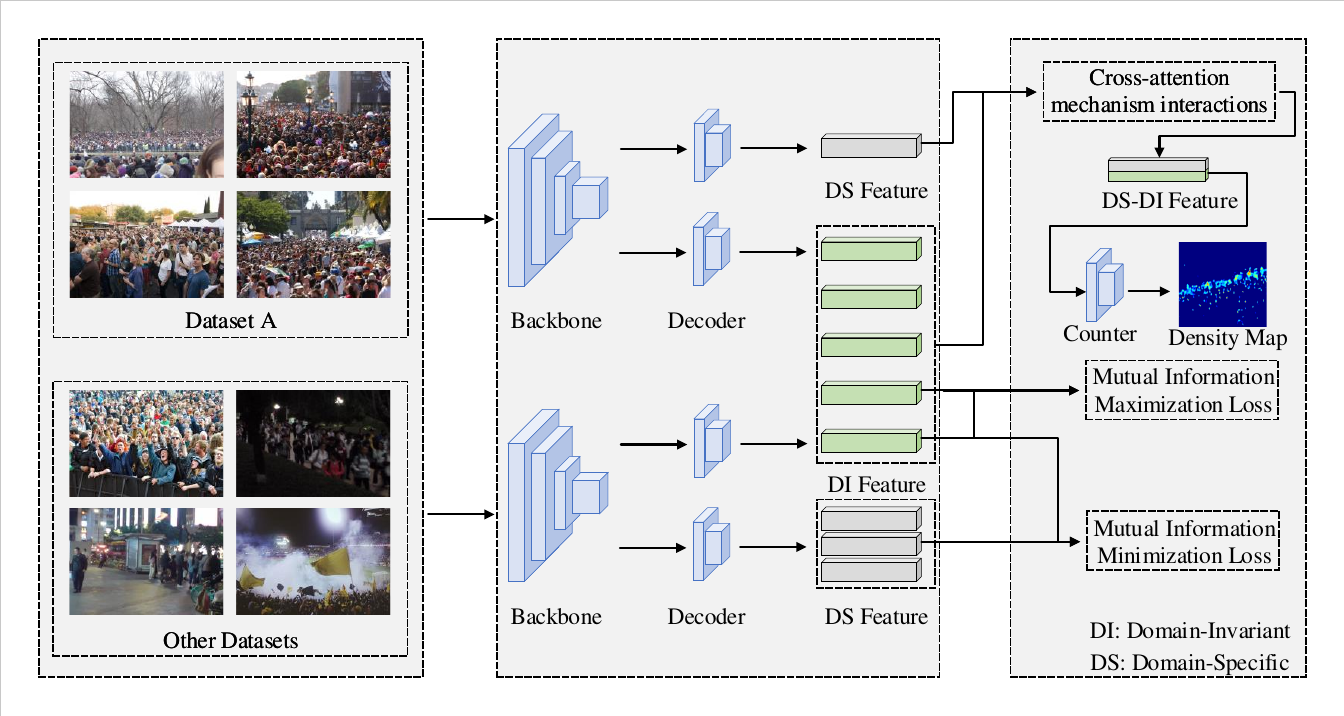}
\end{center}
   \caption{Overview of FSCA-Net. The backbone extracts base features from multiple datasets, which are decomposed into domain-invariant (DI) and domain-specific (DS) features. A cross-attention mechanism enhances inter-dataset interactions, and DI and DS features are further fused into DS-DI features. All features are refined by the decoder to produce the crowd density map. Training is guided by mutual information optimization and a domain-specific strategy, enabling both cross-domain generalization and intra-domain performance.}
\label{fig:framework}
\end{figure*}

The proposed Feature-Separated Cross-Attention Crowd Counting Network (FSCA-Net), illustrated in Figure \ref{fig:framework}, is a multi-dataset framework designed around feature separation, cross-attention interactions, and mutual information optimization. Given inputs from Dataset A and other datasets, the backbone first extracts base features, which are explicitly disentangled into domain-invariant (DI) features and domain-specific (DS) features. At the same time, a cross-attention mechanism is introduced to enhance inter-dataset correlations and facilitate information exchange. The DI and DS features are further fused into DS-DI features, which are passed to the decoder for refinement, ultimately generating the counter density map.

During training, mutual information maximization is employed to improve the generalization of DI features across datasets, while mutual information minimization suppresses redundant associations between DS features and non-target domain information. In addition, a domain-specific training strategy is adopted, where DI features are fused with the DS feature branch of the target domain and supervised using target-domain samples. Through this joint design, FSCA-Net achieves both robust cross-domain generalization and enhanced intra-domain performance.

\subsection{Feature Separation and Fusion}
As illustrated in Figure \ref{fig:framework}, feature separation and fusion constitute the core components of the proposed Feature-Separated Cross-Attention Crowd Counting Network (FSCA-Net), following a separation-then-fusion paradigm. In this design, feature fusion is exclusively realized through a \textbf{cross-attention mechanism}, which first disentangles domain-invariant and domain-specific information, and then strengthens inter-dataset correlations by cross-attention interactions. This framework provides the basis for balancing cross-domain generalization and in-domain counting accuracy. Centered on multi-source image features (Dataset A and other datasets), the process is divided into two sequential stages: feature separation and feature fusion.

\subsubsection{Feature Separation: Decoupling Domain-Invariant and Domain-Specific Features}

Feature separation serves as a critical prerequisite for structured information representation in FSCA-Net. Through a \textbf{feature separation module} (denoted as $\mathcal{S}$), the base features extracted from multi-source images by the backbone are decomposed into two complementary subsets with distinct functions and minimal redundancy. This decoupling reduces interference between shared and dataset-specific information, thus enabling more effective subsequent fusion.

\subsubsection{Mathematical Formulation of Feature Separation}

According to the framework design in Figure \ref{fig:framework}, the feature separation process is formally expressed as:
\begin{equation}
\mathbf{F}_{\text{DI}}, \mathbf{F}_{\text{DS}} = \mathcal{S}(\mathbf{F}_{\text{base}})
\end{equation}
Here, $\mathbf{F}_{\text{base}}$ denotes the base features extracted by the backbone, encompassing low-level visual cues such as textures and spatial structures from Dataset A and other datasets. $\mathbf{F}_{\text{DI}}$ represents the \textbf{Domain-Invariant (DI) Features}, while $\mathbf{F}_{\text{DS}}$ represents the \textbf{Domain-Specific (DS) Features}. These two subsets play distinct yet complementary roles in representation learning:
\begin{itemize}
    \item $\mathbf{F}_{\text{DI}}$: Encodes domain-agnostic, transferable knowledge shared across datasets, such as general human contour patterns and the spatial distribution of dense and sparse crowd regions. These features provide the foundation for the network’s cross-domain generalization.
    \item $\mathbf{F}_{\text{DS}}$: Encodes dataset-specific attributes, such as scene-dependent background textures (e.g., indoor venues vs. outdoor streets) or scale distribution characteristics (e.g., small-scale dense crowds vs. large-scale sparse crowds). These features enhance the model’s capacity to adapt to specific datasets, improving in-domain counting accuracy.
\end{itemize}

\subsubsection{Feature Fusion. Strengthening Multi-Source Feature Correlations via Cross-Attention Mechanism}

Feature fusion constitutes the critical stage following feature separation, aiming to reinforce multi-source feature correlations through \textbf{cross-attention interactions}. As illustrated in Figure \ref{fig:framework}, the cross-attention mechanism is employed as the sole method for implementing feature fusion. By establishing interactive correlations between $\mathbf{F}_{\text{DI}}$ and $\mathbf{F}_{\text{DS}}$ from Dataset A and Other Datasets, this mechanism integrates domain-invariant and domain-specific information, thereby generating fused features with both cross-domain adaptability and in-domain accuracy. These fused representations serve as high-quality inputs for the decoder, ultimately contributing to accurate counter density map estimation.  

Based on the framework in Figure \ref{fig:framework}, the cross-attention mechanism (denoted as $\mathcal{A}_{\text{cross}}$) operates as follows: first, features from Dataset A are assigned as \textbf{Queries (Q)}, while features from Other Datasets serve as \textbf{Keys (K)} and \textbf{Values (V)}. Importantly, queries and keys are drawn from the same feature type (i.e., $\mathbf{Q}_{\text{DI}} = \mathbf{F}_{\text{DI-A}}, \mathbf{K}_{\text{DI}} = \mathbf{F}_{\text{DI-O}}$, or $\mathbf{Q}_{\text{DS}} = \mathbf{F}_{\text{DS-A}}, \mathbf{K}_{\text{DS}} = \mathbf{F}_{\text{DS-O}}$), where the subscript “A” denotes Dataset A and “O” denotes Other Datasets. This design ensures that the interactions capture cross-domain correlations within the same semantic category of features.  

The cross-attention output is computed as:
\begin{equation}
\mathbf{F}_{\text{attn-X}} = \mathcal{A}_{\text{cross}}(\mathbf{Q}_{\text{X}}, \mathbf{K}_{\text{X}}, \mathbf{V}_{\text{X}}) = \text{Softmax}\left( \frac{\mathbf{Q}_{\text{X}} \mathbf{K}_{\text{X}}^T}{\sqrt{d_k}} \right) \mathbf{V}_{\text{X}}
\end{equation}
Here, $\text{X}$ denotes the feature type (either DI or DS), and $d_k$ is the dimensionality of the key features, which scales the dot-product to maintain numerical stability. The Softmax function generates attention weights that emphasize regions with strong inter-dataset correlations (e.g., areas with similar crowd distribution patterns).  

Subsequently, the cross-attention output is fused with the original features of the same type via residual connections to obtain the final fused features:
\begin{equation}
\mathbf{F}_{\text{fusion-X}} = \mathbf{F}_{\text{X-A}} + \mathbf{F}_{\text{attn-X}}
\end{equation}
This process yields fused domain-invariant features $\mathbf{F}_{\text{fusion-DI}}$ and fused domain-specific features $\mathbf{F}_{\text{fusion-DS}}$, thereby completing the fusion process.  

As summarized in Figure \ref{fig:framework}, the cross-attention mechanism fulfills two essential roles:
\begin{enumerate}
    \item For $\mathbf{F}_{\text{DI}}$: Enhances the consistency of domain-invariant information across datasets, improving the generalization ability of DI features.
    \item For $\mathbf{F}_{\text{DS}}$: Mitigates interference from domain-shift-induced variations (e.g., differences in crowd scale distributions) while preserving dataset-specific details, ensuring a balance between specialization and adaptability.
\end{enumerate}
Together, these fused features provide robust support for the decoder, enabling accurate and generalizable crowd counting.  

\subsection{Mutual Information Minimization and Maximization Losses}
To optimize the representational capacity of $\mathbf{F}_{\text{DI}}$ (domain-invariant features) and $\mathbf{F}_{\text{DS}}$ (domain-specific features), Figure~\ref{fig:framework} incorporates two complementary loss functions based on mutual information (MI). Specifically, these losses guide the learning process such that $\mathbf{F}_{\text{DI}}$ attains stronger cross-domain generalization ability, while $\mathbf{F}_{\text{DS}}$ preserves higher domain specificity.

\subsubsection{Mutual Information Maximization Loss (for $\mathbf{F}_{\text{DI}}$ Optimization)}
This loss is specifically designed for optimizing $\mathbf{F}_{\text{DI}}$, with the goal of enhancing cross-domain consistency and facilitating domain adaptation when target domain data is available. Mutual information (MI) measures the statistical dependence between two variables, here $\mathbf{F}_{\text{DI}}$ extracted from the source and target domains, by quantifying how much information one variable provides about the other.

Let $z$ denote the latent representation of $\mathbf{F}_{\text{DI}}$ from the source domain (e.g., $\mathcal{D}_A$, Dataset~A), and $z^{+}$ denote the corresponding latent representation from the target domain (e.g., a subset of $\mathcal{D}_O$, Other Datasets). Both $z$ and $z^{+}$ are obtained by passing $\mathbf{F}_{\text{DI}}$ through a linear embedding, thereby ensuring consistency in the feature space. The MI between $z$ and $z^{+}$ is defined as:
\begin{equation}
I(z; z^{+}) = \mathbb{E}_{p(z, z^+)} \left[ \log \frac{p(z^{+} \mid z)}{p(z^{+})} \right],
\end{equation}
where $p(z, z^{+})$ is the joint distribution of $(z, z^{+})$, $p(z^{+} \mid z)$ is the conditional distribution, and $p(z^{+})$ is the marginal distribution of $z^{+}$. Maximizing $I(z; z^{+})$ enforces $\mathbf{F}_{\text{DI}}$ to capture crowd-related features consistently across domains, thereby supporting effective knowledge transfer.

Since directly estimating $I(z; z^{+})$ is intractable due to unknown probability distributions, we adopt a lower bound approximation guided by a similarity function $f(\cdot, \cdot)$ (e.g., cosine similarity) for stable optimization. Specifically, the following mutual information lower bound ($I_{\text{LB}}$) is used:
\begin{equation}
I_{\text{LB}}(z; z^{+}) = \mathbb{E}\left[ \frac{1}{N} \sum_{i=1}^N 
\log \frac{\exp \left( f(z_i, z_i^{+}) \right)}
{\sum_{j=1}^N \exp \left( f(z_i, z_j^{+}) \right)} \right],
\end{equation}
where $N$ is the number of sampled pairs, and $(z_i, z_i^{+})$ denotes the $i$-th latent representation pair from the source and target domains. By incorporating $I_{\text{LB}}$ as a regularization term into the overall objective, the cross-domain consistency of $\mathbf{F}_{\text{DI}}$ is indirectly reinforced through the maximization of $I_{\text{LB}}$.

\subsubsection{Mutual Information Minimization Loss (for $\mathbf{F}_{\text{DS}}$ Optimization)}

This loss is specifically designed for $\mathbf{F}_{\text{DS}}$ to suppress redundant cross-domain correlations and enforce domain generalization in scenarios where target domain data are unavailable. The primary objective is to ensure that $\mathbf{F}_{\text{DS}}$ from the target domain (e.g., $\mathcal{D}_A$) captures only attributes unique to $\mathcal{D}_A$, while avoiding interference from non-target domains (e.g., $\mathcal{D}_O$).

Direct minimization of mutual information is intractable due to the unknown conditional distribution. To address this, Figure~\ref{fig:framework} adopts the \textit{Contrastive Log-ratio Upper Bound (CLUB)} to estimate an upper bound on the mutual information between $\mathbf{F}_{\text{DS}}$ of the target domain ($z$) and $\mathbf{F}_{\text{DS}}$ of non-target domains ($z^+$). Formally, for latent vectors $z \sim \mathcal{D}_A$ and $z^+ \sim \mathcal{D}_O$, the CLUB upper bound is defined as:
\begin{equation}
\begin{split}
I_{\text{CLUB}}(z; z^{+}) 
&= \mathbb{E}_{p(z, z^+)}\Big[\log p(z^{+} \mid z)\Big] \\
&\quad - \mathbb{E}_{p(z)p(z^+)}\Big[\log p(z^{+} \mid z)\Big],
\end{split}
\end{equation}

where $I_{\text{CLUB}}(z; z^+) \geq I(z; z^+)$ with a non-negative gap $\Delta_1 = I_{\text{CLUB}} - I$. Thus, minimizing $I_{\text{CLUB}}$ indirectly reduces the true mutual information.

Since the conditional distribution $p(z^+ \mid z)$ is inaccessible, a variational distribution $q_\theta(z^+ \mid z)$ (parameterized by $\theta$ and implemented via a lightweight MLP) is employed to approximate it. For $N$ paired samples $\{(z_i, z_i^+)\}_{i=1}^N$, the unbiased estimator of $I_{\text{CLUB}}$ is given as:
\begin{equation}
\begin{split}
\hat{I}_{\text{CLUB}}(z; z^{+}) = \frac{1}{N} \sum_{i=1}^N \Bigg[ & 
\log q_\theta\left(z^{+}_i \mid z_i\right) \\
& - \frac{1}{N} \sum_{j=1}^N \log q_\theta\left(z^{+}_j \mid z_i\right)
\Bigg]
\end{split}
\end{equation}

To ensure $\hat{I}_{\text{CLUB}}$ reliably approximates the true upper bound, the estimation error 
\begin{equation}
\begin{split}
\Delta_2 &= \text{KLD}\!\left(p(z^+, z) \parallel q_\theta(z^+, z)\right) \\
&= \mathbb{E}_{p(z, z^+)}\!\left[\log p(z^+ \mid z)\right] 
   - \mathbb{E}_{p(z, z^+)}\!\left[\log q_\theta(z^+ \mid z)\right]
\end{split}
\end{equation}

must be minimized. Since only the second term depends on $q_\theta$, this is achieved by minimizing the negative log-likelihood loss:
\begin{equation}
L_{\Delta_2} = -\frac{1}{N} \sum_{i=1}^N \log q_\theta(z_i^{+} \mid z_i).
\end{equation}

During training, an alternating optimization strategy is employed: 
(1) $L_{\Delta_2}$ updates $q_\theta$ to better approximate $p(z^+ \mid z)$, 
while (2) $\hat{I}_{\text{CLUB}}$ guides the suppression of mutual information between $\mathbf{F}_{\text{DS}}$ and $\mathbf{F}_{\text{DI}}$.

In summary, minimizing $\hat{I}_{\text{CLUB}}$ reduces redundancy among domain-specific features, thereby encouraging feature diversity and improving robustness under domain generalization.

\subsection{Counting Network and Final Result Generation}
The counting network of the Feature-Separated Cross-Attention Crowd Counting Network (FSCA-Net) primarily comprises a counter. Its core function is to transform the optimized features ($\mathbf{F}_{\text{DS-DI}}$) into accurate crowd count predictions, serving as the terminal module within the framework illustrated in Figure \ref{fig:framework}.

\subsubsection{Crowd Count Calculation and Domain-Specific Training}
The total crowd count $\hat{N}$ for a single input image is obtained by summing over all pixels of the predicted density map:
\begin{equation}
\hat{N} = \sum_{i=1}^H \sum_{j=1}^W \hat{y}_{i,j}.
\end{equation}

To enhance adaptation to dataset-specific characteristics, a domain-specific training strategy is employed, as illustrated in Figure \ref{fig:framework}. For a target domain (e.g., $\mathcal{D}_A$, Dataset A), the cross-domain generalizable features $\mathbf{F}_{\text{DI}}$ are fused with the domain-specific features $\mathbf{F}_{\text{DS}}$ of the target domain branch. Supervised learning is then performed on the fused features using labeled samples from $\mathcal{D}_A$. This approach ensures that the network retains generalizable knowledge through $\mathbf{F}_{\text{DI}}$ while simultaneously capturing target-domain-specific patterns via $\mathbf{F}_{\text{DS}}$, thereby achieving a balance between cross-domain generalization and in-domain counting accuracy.

The final loss function $ L_{ {count }}$ that we use to optimize the counting network is given by:

\begin{equation}
L_{ {count }}=L_{ {counter }}+\beta_1 I(z; z^{+})+\beta_2 L_{\Delta_2}
\end{equation}

\noindent where $L_{ {counter }}$ is the counting loss function of mainstream counting methods \cite{CSRNet, BL, DM-Count, MAN}, $I(z; z^{+})$ is the mutual information minimization loss. Minimizing $L_{\Delta_2}$ could make sure that $\hat{{I}}_{{CLUB}}$ can hold a MI upper bound. We evaluated the model performance via cross-validation for $\beta_1 \in \{0.01,0.1,1\}$ and $\beta_2 \in \{0.001,0.01,0.1\}$. The results show that $\beta_1=0.1$ and $\beta_2=0.01$ achieve the optimal balance among the different losses in the final loss function.

As shown in Figure \ref{fig:framework}, FSCA-Net establishes a closed-loop optimization pipeline: multi-source images are first processed by the backbone network and cross-attention mechanism to produce enhanced features; these features are subsequently separated into $\mathbf{F}_{\text{DI}}$ and $\mathbf{F}_{\text{DS}}$, with each subset optimized via mutual information-based loss functions; finally, the decoder integrates the optimized features to generate density maps and corresponding crowd counts. By jointly leveraging feature separation, cross-attention interactions, and mutual information optimization, FSCA-Net effectively addresses critical challenges in multi-dataset crowd counting, such as domain shift and limited generalization, providing more accurate and robust crowd counting results across unseen scenarios.

\section{Experiments}

In this section, we present experimental results to evaluate the effectiveness of our proposed methods. We begin by describing the experimental setups in detail, including the datasets used, network architecture, training procedures, and evaluation metrics. Then, we compare our approach with recent state-of-the-art methods. Finally, we conduct ablation studies to further demonstrate the efficacy of our proposed methods.

\subsection{Experimental Setups}

\noindent
\textbf{Dataset.} Experiments are conducted on several widely used crowd counting datasets, including ShanghaiTech A (SHA) \cite{MCNN}, ShanghaiTech B (SHB) \cite{MCNN}, UCF-QNRF (QNRF) \cite{UCF-QNRF}, and JHU-CROWD++ (JHU) \cite{JHU2}. SHA comprises 482 high-density images collected from the Internet, whereas SHB consists of images captured by surveillance cameras in busy streets, resulting in relatively sparser crowd scenes compared to SHA. QNRF is a large-scale dataset featuring diverse crowd scenarios, predominantly with dense crowds. JHU is a comprehensive dataset containing 4,372 images and 1.51 million annotations. Compared with existing datasets, our selected datasets cover a wide range of scenarios and environmental conditions and provide richer annotations, including dot maps, approximate bounding boxes, and blur levels.

\noindent
\textbf{Network Architecture and Training Details.} We adopt the BL \cite{BL}, CSRNet \cite{CSRNet}, DM-Count \cite{DM-Count} and MAN \cite{MAN} as the baseline for our proposed methods, and maintain training settings consistent with those reported in BL \cite{BL} to ensure fair comparisons. e adopt three-layer convolutional regressors as the decoder, $Conv_{512\times256\times3\times3}$ +
$Conv_{256\times256\times1\times1}$ + $Conv_{256\times512\times3\times3}$ (Input Channels $\times$ Output Channels $\times$ Kernel Height $\times$ Kernel Width).Specifically, we adopt three-layer convolutional regressors as the counter, $Conv_{512\times256\times3\times3}$ +
$Conv_{256\times128\times3\times3}$ + $Conv_{128\times1\times3\times3}$ (Input Channels $\times$ Output Channels $\times$ Kernel Height $\times$ Kernel Width).

\noindent
\textbf{Evaluation Metrics.} The performance of the proposed method is evaluated using the widely adopted Mean Absolute Error (MAE) and Root Mean Squared Error (MSE) metrics, defined as:
\begin{equation}
MAE = \frac{1}{N} \sum_{i=1}^{N} |\hat{C}_{i} - C_{i}|, \quad
MSE = \sqrt{\frac{1}{N} \sum_{i=1}^{N} (\hat{C}_{i} - C_{i})^2},
\end{equation}
where $N$ denotes the number of test images, $\hat{C}_{i}$ is the estimated crowd count obtained from the predicted density map, and $C_{i}$ is the corresponding ground truth count.

\subsection{Experimental Comparison}

\noindent\textbf{Comparison with Baselines.}

We evaluate the proposed FSCA-Net by integrating it with four representative state-of-the-art crowd counting methods: BL \cite{BL}, CSRNet \cite{CSRNet}, DM-Count \cite{DM-Count}, and MAN \cite{MAN}. As reported in Table~\ref{tab:method_comparison}, for all baseline models, training on multiple datasets simultaneously (denoted as (M)) leads to inferior performance compared to training on each dataset individually (denoted as (S)), despite the increased availability of training samples. This performance degradation arises because a single counting model is unable to effectively accommodate the large scale variations and distributional discrepancies inherent across datasets, resulting in underfitting and poor generalization.

In contrast, when equipped with our FSCA-Net, these baseline models exhibit substantial improvements under multi-dataset training. For example, FSCA-Net+CSRNet(M) and FSCA-Net+BL(M) not only outperform their corresponding multi-dataset baselines but also surpass the single-dataset baselines. These results demonstrate that explicitly disentangling and re-organizing features through FSCA-Net enables more effective knowledge transfer across datasets while simultaneously enhancing dataset-specific adaptation.

\begin{table*}[htbp]
  \centering
  \caption{Counting performance comparisons with baseline methods. (S) indicates that the model is trained on each dataset separately. (M) indicates that the model is jointly trained on multiple datasets.}
  \begin{tabularx}{\textwidth}{l*{8}{l}}
    \toprule
    \multirow{2}{*}{Method} & \multicolumn{2}{c}{UCF-QNRF} & \multicolumn{2}{c}{Shanghai\_A} & \multicolumn{2}{c}{JHU-Crowd++} & \multicolumn{2}{c}{UCF\_CC\_50} \\
    \cmidrule(lr){2-3} \cmidrule(lr){4-5} \cmidrule(lr){6-7} \cmidrule(lr){8-9}
    & MAE & MSE & MAE & MSE & MAE & MSE & MAE & MSE \\
    \midrule
    CSRNET(S) \cite{CSRNet} & 110.6 & 190.1 & 68.6 & 115.0 & 85.9 & 309.2 & 266.1 & 397.5 \\
    CSRNET(M) & 121.5 & 168.2 & 70.4 & 123.1 & 94.8 & 282.7 & 332.6 & 394.5 \\ 
    \textbf{FSCA-Net+CSRNET(M)} & \textbf{87.6\textsuperscript{\textcolor{red}{(-27.9\%)}}} & \textbf{134.4\textsuperscript{\textcolor{red}{(-20.1\%)}}} & \textbf{46.0\textsuperscript{\textcolor{red}{(-34.7\%)}}} & \textbf{87.6\textsuperscript{\textcolor{red}{(-28.8\%)}}} & \textbf{56.2\textsuperscript{\textcolor{red}{(-40.7\%)}}} & \textbf{231.6\textsuperscript{\textcolor{red}{(-18.1\%)}}} & \textbf{156.9\textsuperscript{\textcolor{red}{(-52.9\%)}}} & \textbf{231.4\textsuperscript{\textcolor{red}{(-41.3\%)}}} \\
    \midrule
    BL(S) \cite{BL} & 88.7 & 154.8 & 68.4 & 97.8 & 67.1 & 268.9 & 229.3 & 308.1 \\
    BL(M) & 92.2 & 158.6 & 70.6 & 99.7 & 69.4 & 276.2 & 235.0 & 317.3 \\
    \textbf{FSCA-Net+BL(M)} & \textbf{83.8\textsuperscript{\textcolor{red}{(-9.1\%)}}} & \textbf{144.5\textsuperscript{\textcolor{red}{(-8.9\%)}}} & \textbf{46.1\textsuperscript{\textcolor{red}{(-34.7\%)}}} & \textbf{78.1\textsuperscript{\textcolor{red}{(-21.7\%)}}} & \textbf{58.4\textsuperscript{\textcolor{red}{(-15.8\%)}}} & \textbf{237.1\textsuperscript{\textcolor{red}{(-14.2\%)}}} & \textbf{230.2\textsuperscript{\textcolor{red}{(-2.0\%)}}} & \textbf{274.7\textsuperscript{\textcolor{red}{(-13.4\%)}}} \\
    \midrule
    DM(S) \cite{DM-Count} & 85.6 & 148.3 & 59.7 & 95.7 & 66.0 & 261.4 & 211.0 & 291.5 \\
    DM(M) & 105.7 & 167.3 & 65.2 & 110.1 & 62.8 & 234.2 & 271.2 & 403.4 \\
    \textbf{FSCA-Net+DM(M)} & \textbf{75.1\textsuperscript{\textcolor{red}{(-29.0\%)}}} & \textbf{135.0\textsuperscript{\textcolor{red}{(-19.3\%)}}} & \textbf{49.9\textsuperscript{\textcolor{red}{(-23.5\%)}}} & \textbf{85.3\textsuperscript{\textcolor{red}{(-22.6\%)}}} & \textbf{54.7\textsuperscript{\textcolor{red}{(-12.9\%)}}} & \textbf{229.0\textsuperscript{\textcolor{red}{(-2.2\%)}}} & \textbf{181.7\textsuperscript{\textcolor{red}{(-33.0\%)}}} & \textbf{242.9\textsuperscript{\textcolor{red}{(-39.8\%)}}} \\
    \midrule
    MAN(S) \cite{MAN}  & 77.3 & 131.5 & 56.8 & 90.3 & 53.4 & 209.9 & - & - \\
    MAN(M) & 83.2 & 142.0 & 61.8 & 97.5 & 57.6 & 226.6 & - & - \\
    \textbf{FSCA-Net+MAN(M)} & \textbf{73.5\textsuperscript{\textcolor{red}{(-11.7\%)}}} & \textbf{124.5\textsuperscript{\textcolor{red}{(-12.3\%)}}} & \textbf{49.3\textsuperscript{\textcolor{red}{(-20.3\%)}}} & \textbf{78.2\textsuperscript{\textcolor{red}{(-19.8\%)}}} & \textbf{46.0\textsuperscript{\textcolor{red}{(-20.1\%)}}} & \textbf{192.4\textsuperscript{\textcolor{red}{(-15.1\%)}}} & - & - \\
    \bottomrule
  \end{tabularx}
  \label{tab:method_comparison}
\end{table*}

\noindent\textbf{Comparison with State of The Arts.}

To validate the effectiveness of the proposed Feature-Separated Cross-Attention Crowd Counting Network (FSCA-Net), which addresses occlusion, scale variation, and cross-dataset distributional shifts via explicit separation of domain-invariant (DI) and domain-specific (DS) features, coupled with cross-attention interactions and mutual information optimization, we conduct extensive comparative experiments against state-of-the-art methods on four representative crowd counting benchmarks, namely UCF-QNRF, Shanghai\_A, JHU-Crowd++, and UCF\_CC\_50. Quantitative results, detailed in Table \ref{tab:state_of_the_art_comparison}, demonstrate the superiority of FSCA-Net in both cross-domain generalization and intra-domain performance. Notably, FSCA-Net achieves leading counting accuracy across multiple datasets: FSCA-Net+CSRNET(M) delivers the lowest MAE (\textcolor{red}{46.0}) and competitive MSE (\textcolor{blue}{87.6}) on Shanghai\_A, while also securing the best MAE (\textcolor{red}{156.9}) and MSE (\textcolor{red}{231.4}) on UCF\_CC\_50; FSCA-Net+MAN(M) further sets new benchmarks on JHU-Crowd++ with the lowest MAE (\textcolor{red}{46.0}) and MSE (\textcolor{red}{192.4}), and additionally achieves the best MSE (\textcolor{red}{124.5}) on UCF-QNRF and the lowest MSE (\textcolor{red}{78.2}) on Shanghai\_A. 

Importantly, all these performance gains are realized using universal models trained jointly on multiple datasets, eliminating the need for dataset-specific model customization—a critical advantage for real-world deployment. Equally compelling is FSCA-Net’s ability to serve as an effective plug-and-play enhancement for existing baseline methods: when integrated with conventional models such as CSRNet, BL, and DM-Count, the resulting FSCA-Net variants (FSCA-Net+CSRNET(M), FSCA-Net+BL(M), FSCA-Net+DM(M)) consistently reduce MAE and MSE across datasets. For example, FSCA-Net+CSRNET(M) reduces the MAE of the original CSRNet on Shanghai\_A from 68.2 to \textcolor{red}{46.0} and on UCF\_CC\_50 from 266.1 to \textcolor{red}{156.9}, while FSCA-Net+DM(M) lowers the MAE of DM-Count on JHU-Crowd++ from 66.0 to 54.7. These results collectively confirm that FSCA-Net’s design—leveraging disentangled DI/DS features to preserve transferable knowledge and retain domain-specific properties—effectively mitigates cross-dataset distribution conflicts and enhances representation adaptability, thereby outperforming existing methods and validating its value as a generalizable enhancement for crowd counting tasks.

\begin{table*}[htbp]
  \centering
  \caption{Counting performance comparisons with state-of-the-art methods. \textcolor{red}{RED} indicates the best performance and \textcolor{blue}{BLUE} indicates the second-best. Lower values indicate better performance.}
  \begin{tabularx}{\textwidth}{l*{8}{>{\centering\arraybackslash}X}}
    \toprule
    \multirow{2}{*}{Method} & \multicolumn{2}{c}{UCF-QNRF} & \multicolumn{2}{c}{Shanghai\_A} & \multicolumn{2}{c}{JHU-Crowd++} & \multicolumn{2}{c}{UCF\_CC\_50} \\
    \cmidrule(lr){2-3} \cmidrule(lr){4-5} \cmidrule(lr){6-7} \cmidrule(lr){8-9}
    & MAE & MSE & MAE & MSE & MAE & MSE & MAE & MSE \\
    \midrule
    ASNPE \cite{58} & 89.1 & 159.7 & 57.8 & 97.1 & - & - & 174.8 & 251.6 \\
    RPNNet \cite{58} & - & - & 61.2 & 96.9 & - & - & - & - \\
    MNA \cite{45} & 85.8 & 150.6 & 61.9 & 99.6 & 67.7 & 258.5 & - & - \\
    ADSCNET \cite{2} & {71.3} & {132.5} & 55.4 & 97.7 & - & - & 198.4 & 267.3 \\
    CSRNet \cite{CSRNet} & 110.6 & 190.1 & 68.2 & 115.0 & 85.9 & 309.2 & 266.1 & 397.5 \\
    BL \cite{BL} & 88.7 & 154.8 & 68.4 & 101.8 & 67.1 & 268.9 & 229.3 & 308.1 \\
    M-SFANET \cite{M-SFANet} & 85.6 & 151.2 & 59.7 & 95.7 & 65.5 & 265.4 & 212.3 & 291.5 \\
    DM \cite{DM-Count} & 85.6 & 148.3 & 59.7 & 95.7 & 66.0 & 261.7 & {161.0} & 276.8 \\
    GL \cite{GL} & 84.3 & 147.5 & 61.3 & 95.4 & - & - & - & - \\
    ChfL \cite{ChfL} & {80.3} & {137.6} & 57.5 & 94.3 & - & - & - & - \\
    GauNet \cite{GauNet} & 81.6 & 153.7 & {54.8} & {89.1} & - & - & 186.3 & \textcolor{blue}{256.5} \\
    PET \cite{liu2023point} & 79.5 & 144.3 & 47.4 & 75.0 & 58.5 & 238.0 & - & - \\
    CrowdHat \cite{wu2023boosting} & 75.1 & 126.7 & 51.2 & 81.9 & 52.3 & 211.8 & - & - \\
    STEERER \cite{STEERER} & 74.3 & 128.3 & 54.5 & 86.9 & 54.3 & 238.3 & - & - \\
    APGCC \cite{chen2024improving} & 80.0 & 136.6 & 48.8 & 76.7 & 54.3 & 255.9 & - & - \\
    CrowDiff \cite{CrowDiff} & \textcolor{red}{68.9} & \textcolor{blue}{125.6} & 47.4 &  \textcolor{red}{75.0} & \textcolor{blue}{47.3} & \textcolor{blue}{198.9} & - & - \\
    P2R \cite{P2R} & 83.3 & 138.1 & 51.0 & 79.7 & 58.8 & 253.1 & - & - \\
    \midrule
    FSCA-Net+CSRNET(M) & 87.6 & 134.4 & \textcolor{red}{46.0} & {87.6} & 56.2 & 231.6 & \textcolor{red}{156.9} & \textcolor{red}{231.4} \\
    FSCA-Net+BL(M) & 83.8 & 144.5 & \textcolor{blue}{46.1} & \textcolor{blue}{78.1} & 58.4 & 237.1 & 230.2 & 274.7 \\
    FSCA-Net+DM(M) & 75.1 & 135.0 & 49.9 & 85.3 & 54.7 & {229.0} & 181.7 & 242.9 \\
    FSCA-Net+MAN(M) & \textcolor{blue}{73.5} & \textcolor{red}{124.5} & 49.3 &{78.2} & \textcolor{red}{46.0} & \textcolor{red}{192.4} & - & - \\
    \bottomrule
  \end{tabularx}
  \label{tab:state_of_the_art_comparison}
\end{table*}

\subsection{Ablation Study}
To evaluate the independent contributions and synergistic effects of the core components in FSCA-Net—including the Feature Separation Module (FSM), Cross-Attention Feature Fusion Module (CAFFM), Mutual Information Minimization Loss (MIML), and Mutual Information Maximization Loss (MAXL)—we conducted a systematic ablation study using the FSCA-Net+BL(M) variant as the baseline, with all experiments performed on the UCF-QNRF dataset. The design of each ablation setting adheres to a "component removal" principle to isolate the impact of individual modules: removing the FSM means abandoning the explicit disentanglement of domain-invariant (DI) and domain-specific (DS) features, and instead using the original BL(M) model’s feature extraction pipeline; removing the CAFFM replaces the cross-attention-based fusion of DI and DS features with a simple concatenation (concat) operation, which only aggregates feature channels without capturing inter-feature correlations; removing MIML or MAXL involves omitting the corresponding mutual information constraint, thereby disabling the optimization of feature discrimination (for MIML) or cross-domain generalization (for MAXL), respectively.  

Table \ref{tab:ablation} presents the quantitative results of the ablation study, with MAE and MSE as the evaluation metrics (lower values indicate better performance). The full FSCA-Net+BL(M) model (with all components enabled, i.e., FSM $\checkmark$, CAFFM $\checkmark$, MIML $\checkmark$, MAXL $\checkmark$) achieves the optimal performance, with an MAE of 83.8 and MSE of 144.5. This confirms that the synergistic effect of all core components is critical to maximizing the model’s counting accuracy.  

When the FSM is removed (the second row in Table \ref{tab:ablation}), the model’s performance degrades significantly: MAE increases by 4.9 (from 83.8 to 88.7) and MSE rises by 10.3 (from 144.5 to 154.8). This degradation demonstrates that the FSM—by explicitly disentangling DI features (which capture cross-domain generalizable semantic information) and DS features (which encode dataset-specific characteristics like background texture and perspective)—effectively mitigates the distributional conflict between features from different domains, a capability that the original model (without FSM) lacks.  

Omitting the CAFFM (the third row) also leads to noticeable performance loss: MAE increases by 1.2 (to 85.0) and MSE increases by 3.7 (to 148.2) compared to the full model. This result highlights the limitations of simple feature concatenation: unlike the CAFFM, which leverages cross-attention to model fine-grained interactions between DI and DS features and emphasize discriminative information, concat only performs channel-wise aggregation without optimizing feature relevance—ultimately reducing the quality of fused features for density map regression.  

Removing MIML (the fourth row) or MAXL (the fifth row) similarly impairs model performance. Without MIML, MAE rises to 86.8 and MSE to 150.6; without MAXL, MAE reaches 87.2 and MSE remains at 148.8. MIML is designed to minimize redundant correlations between DS features and non-target domain information, ensuring that DS features only encode dataset-specific properties; its removal leads to "noisy" DS features that interfere with cross-domain adaptation. MAXL, by contrast, maximizes the generalization of DI features across datasets, and its omission weakens the model’s ability to transfer knowledge between domains—both resulting in reduced counting accuracy.  

Collectively, these ablation results validate the necessity of each core component in FSCA-Net: the FSM lays the foundation for effective feature disentanglement, the CAFFM optimizes the fusion of disentangled features, and MIML/MAXL jointly refine feature discrimination and generalization. Their synergistic integration is key to the superior performance of FSCA-Net, and no single component can be replaced or omitted without causing performance degradation.

\begin{table}[htbp]
    \centering
    \caption{Ablation study using FSCA-Net+BL(M) on UCF-QNRF dataset. Removing the FSM means using the original model. Removing the CAFFM means using a simple concatenation (concat) operation between domain-invariant (DI) features and domain-specific (DS) features.}
    \begin{tabular}{cccc|c}
        \toprule
        FSM & CAFFM & MIML & MAXL & Result (MAE/MSE)\\
        \midrule
        $\checkmark$ & $\checkmark$ & $\checkmark$ & $\checkmark$ & 83.8 / 144.5\\
        & $\checkmark$ & $\checkmark$ & $\checkmark$ & 88.7 / 154.8\\
        $\checkmark$ & & $\checkmark$ & $\checkmark$ & 85.0 / 148.2\\
        $\checkmark$ & $\checkmark$ & & $\checkmark$ & 86.8 / 150.6\\
        $\checkmark$ & $\checkmark$ & $\checkmark$ & & 87.2 / 148.8\\
        \bottomrule
    \end{tabular}
    \label{tab:ablation}
\end{table}

\subsection{Generalization to Unseen Datasets}

To further illustrate that FSCA-Net enables counting models to better generalize to unseen scenes—leveraging only \textbf{domain-invariant (DI) features} for cross-dataset adaptation—we conduct a cross-dataset evaluation. In this experiment, FSCA-Net (focusing on DI feature extraction) and its integrated baseline models are exclusively trained on the UCF-QNRF dataset, with no additional training data or parameter fine-tuning on the target datasets; the models are directly evaluated on three unseen benchmark datasets (Shanghai\_A, JHU-Crowd++, and UCF\_CC\_50). Quantitative results of this evaluation are presented in Table \ref{tab:generalization_unseen}.  

As shown in the table, FSCA-Net consistently enhances the cross-dataset generalization performance of all baseline models. For instance, when integrated with CSRNet, FSCA-Net reduces the MAE of CSRNet by 16.9\% on Shanghai\_A, 18.5\% on JHU-Crowd++, and 37.0\% on UCF\_CC\_50; even for the relatively robust MAN baseline, FSCA-Net still achieves a 22.4\% MAE reduction on Shanghai\_A and a 24.7\% MAE reduction on UCF\_CC\_50. These improvements confirm that the DI features extracted by FSCA-Net—which capture cross-domain generalizable semantic and structural information (while excluding dataset-specific noise like background textures or illumination variations)—effectively mitigate distributional shifts between the source (UCF-QNRF) and unseen target datasets.  

Notably, FSCA-Net’s reliance on only DI features for generalization avoids the need for domain-specific feature adaptation or target-domain data participation during training, making it a lightweight and practical solution for cross-dataset crowd counting. Compared with baseline models that use mixed feature representations (combining domain-specific and invariant information), FSCA-Net’s focus on DI features ensures more stable and transferable performance across unseen scenes, ultimately achieving the best overall cross-dataset evaluation results among all compared methods.

\begin{table}[htbp]
  \centering
  \caption{Generalization to unseen datasets. The model is trained on UCF-QNRF while tested on the other datasets. Improvement is calculated as percentage reduction relative to baseline methods.}
  \adjustbox{max width=1.0\linewidth}{
  \begin{tabular*}{\linewidth}{@{\extracolsep{\fill}} lrrrrrr @{}}
    \toprule
    \multirow{2}{*}{UCF-QNRF $\rightarrow$} & \multicolumn{2}{c}{Shanghai\_A} & \multicolumn{2}{c}{JHU-Crowd++} & \multicolumn{2}{c}{UCF\_CC\_50} \\
    \cmidrule(lr){2-3} \cmidrule(lr){4-5} \cmidrule(lr){6-7}
    Method & MAE & MSE & MAE & MSE & MAE & MSE \\
    \midrule
    L2SM \cite{56} & 73.4 & 119.4 & - & - & - & - \\
    S-DCNET \cite{55} & 61.8 & 102.8 & - & - & - & - \\
    \midrule
    CSRNET \cite{CSRNet} & 75.3 & 138.7 & 91.4 & 317.0 & 389.8 & 659.6 \\
    FSCA-Net+CSRNET & 62.6 & 100.5 & 74.5 & 268.7 & 245.6 & 398.2 \\ 
    Improvement (\%) & \textcolor{red}{16.9} & \textcolor{red}{27.5} & \textcolor{red}{18.5} & \textcolor{red}{15.2} & \textcolor{red}{37.0} & \textcolor{red}{39.6} \\
    \midrule
    BL \cite{BL} & 69.8 & 123.8 & 81.2 & 303.8 & 309.6 & 537.1 \\
    FSCA-Net+BL & 56.0 & 91.4 & 70.8 & 264.0 & 226.7 & 328.5 \\ 
    Improvement (\%) & \textcolor{red}{19.8} & \textcolor{red}{26.2} & \textcolor{red}{12.8} & \textcolor{red}{13.1} & \textcolor{red}{26.8} & \textcolor{red}{38.8} \\
    \midrule
    DM \cite{DM-Count} & 69.3 & 120.6 & 85.2 & 303.4 & 317.8 & 550.2 \\
    FSCA-Net+DM & 55.0 & 92.5 & 73.9 & 266.2 & 243.5 & 360.1 \\
    Improvement (\%) & \textcolor{red}{20.6} & \textcolor{red}{23.3} & \textcolor{red}{13.3} & \textcolor{red}{12.3} & \textcolor{red}{23.4} & \textcolor{red}{34.6} \\
    \midrule
    MAN \cite{MAN} & 67.1 & 122.1 & 79.0 & 282.1 & 295.3 & 512.7 \\
    FSCA-Net+MAN & 52.1 & 90.6 & 71.0 & 252.9 & 222.4 & 337.2 \\
    Improvement (\%) & \textcolor{red}{22.4} & \textcolor{red}{25.8} & \textcolor{red}{10.1} & \textcolor{red}{10.3} & \textcolor{red}{24.7} & \textcolor{red}{34.2} \\
    \bottomrule
  \end{tabular*}}
  \label{tab:generalization_unseen}
\end{table}

\begin{figure*}[htbp]
    \begin{center}
    \includegraphics[width=1.0\linewidth]{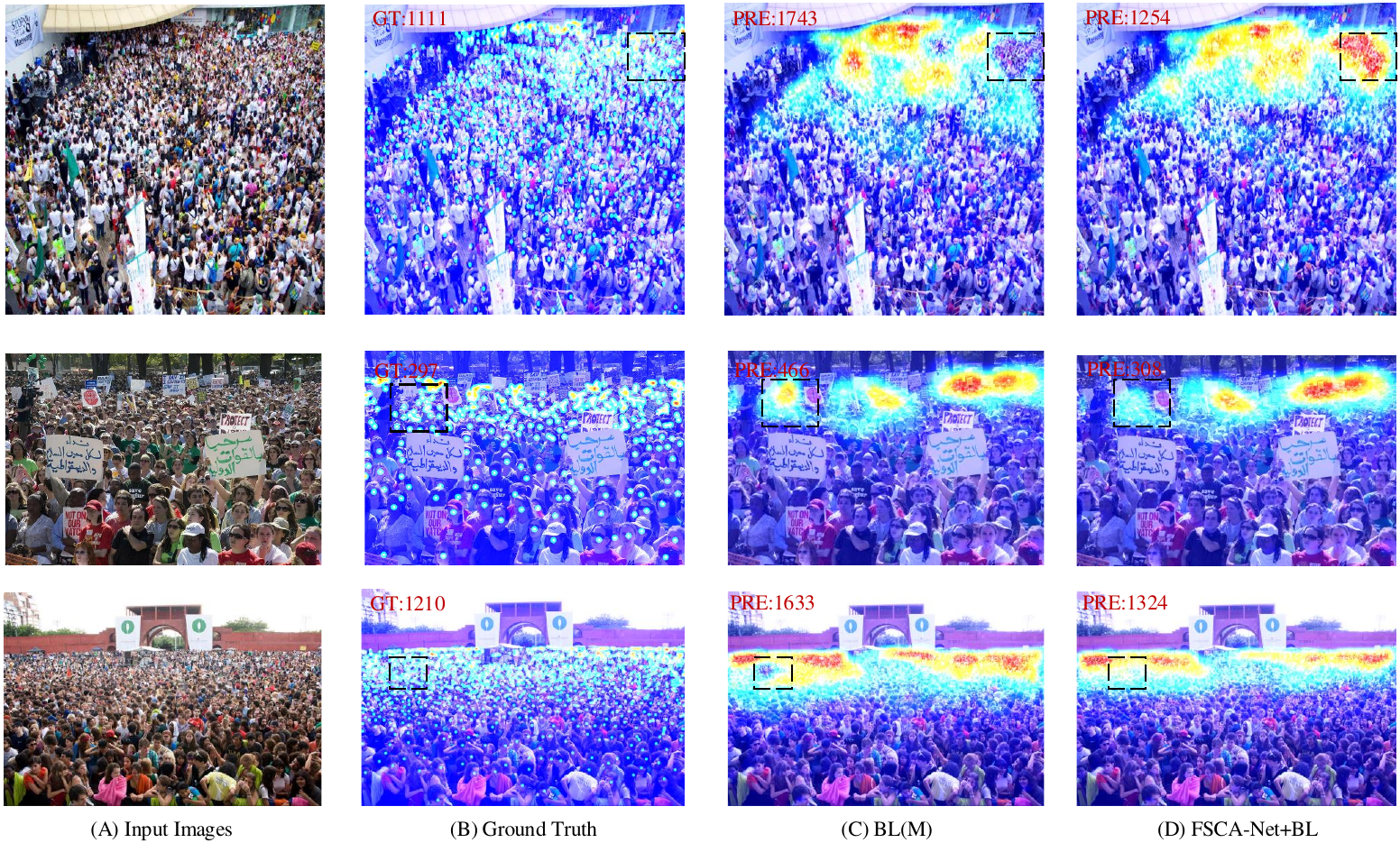}
    \end{center}
   \caption{Visual comparison of crowd counting and density prediction results. (A) Input Images; (B) Ground Truth; (C) BL (M); (D) FSCA-Net+BL.}
    \label{fig:density}
    \end{figure*}

\begin{figure}[htbp]
    \begin{center}
    \includegraphics[width=1.0\linewidth]{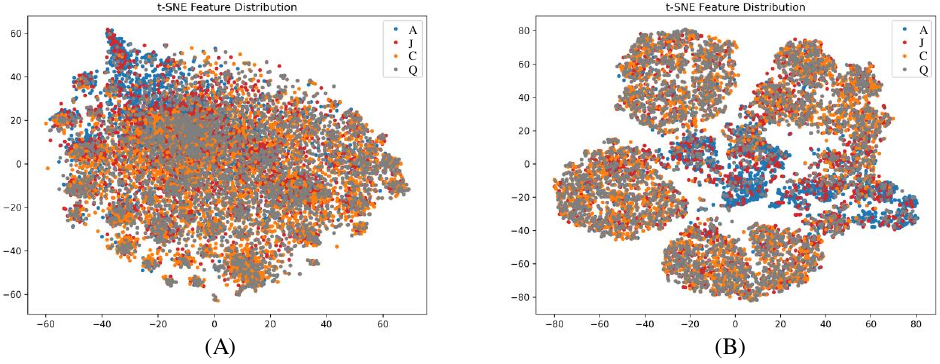}
    \end{center}
   \caption{t-SNE visualization of feature distributions: (A) features from a jointly trained original method (unable to separate domain-invariant and domain-specific features); (B) features extracted by the proposed FSCA-Net.}
    \label{fig:tsne}
    \end{figure}

\subsection{Complexity analysis}
To evaluate the complexity of our method, we present a comparison of parameters, FLOPs, inference speed, and counting results on the ShanghaiTech PartA dataset in Table \ref{tab6}. All experiments are conducted in the same environment: parameters and FLOPs are calculated with an input size of 512×512 on a single NVIDIA 3090 GPU, and the inference time is the average of 100 runs for testing a 1024×768 image sample, in line with the ShanghaiTech A benchmark \cite{MCNN}.
As shown in Table \ref{tab6}, while our FSCA-Net+CSRNet model has slightly increased parameters (20.60 M vs. CSRNet's 16.26 M) and a marginally longer inference time (0.042 s vs. CSRNet's 0.038 s), it achieves a significantly better MSE of 87.6 in PartA counting. Notably, with an inference speed enabling processing of around 23.8 images per second (1/0.042 = 23.8), our model still meets real-time application requirements in practical crowd counting scenarios.

\begin{table*}[htbp]
\caption{Comparison of the Parameters (M), FLOPs (G), Inference speed (s / 100 images) and results on the PartA dataset \cite{MCNN}. Note: The parameters and FLOPs are computed with the input size of 512×512 on a single NVIDIA 3090 GPU. The inference time is the average time of 100 runs on testing a 1024×768 sample.}
\centering
\begin{tabular*}{\textwidth}{@{\extracolsep{\fill}}llllll@{}}
\toprule
Method & Backbone & Parameters(M) & FLOPs(G) & Inference speed(s) & MSE In PartA\\
\midrule
PCC-Net\cite{TCSVT3} & FS & $0.51$ & $43.87$ & $0.013$ & 124.0 \\
CAN\cite{CAN} & VGG16 & $18.10$ & $114.83$ & $0.047$ & 100.0\\
SCAR\cite{SCAR} & VGG16 & $16.29$ & $108.44$ & $0.047$ & 110.0\\
M-SFANet\cite{M-SFANet} & VGG16 & $22.88$ & $115.14$ & $0.058$ & $94.5$ \\
SFCN\cite{SFCN} & ResNet101 & $38.60$ & $162.03$ & $0.096$ & 104.5 \\
\midrule
CSRNet\cite{CSRNet} & VGG16 & $16.26$ & $108.34$ & $0.038$ & 115.0 \\
FSCA-Net+CSRNet & VGG16 & $20.60$ & $124.64$ & $0.042$ & 87.6 \\
\bottomrule
\end{tabular*}
\label{tab6}
\end{table*}

\subsection{Visualization}
To demonstrate that the proposed FSCA-Net can effectively extract domain-invariant features, we analyze the t-SNE feature distribution visualizations of the jointly trained original method (Figure \ref{fig:tsne} (A)) and FSCA-Net (Figure \ref{fig:tsne} (B)).

In Figure \ref{fig:tsne} (A), representing the original method that fails to separate domain-invariant and domain-specific related features, the points of different colors (corresponding to different domains or feature types) are highly mixed and clustered together, especially with a dense agglomeration in the central area. This indicates that the features learned by the ordinary method are entangled, containing both domain-invariant and domain-specific information in an unstructured manner, making it difficult to distinguish and utilize domain-invariant features alone.

In contrast, Figure \ref{fig:tsne} (B) for FSCA-Net shows a much more organized and dispersed distribution of points. The features of different colors are better separated, and there is no excessive central clustering as seen in Figure \ref{fig:tsne} (A). The domain-invariant features (which should be shared across domains and thus show more consistent distribution patterns) can be observed to form clearer, more distinct clusters that are less intertwined with domain-specific features. This improved separation and dispersion of feature points demonstrate that FSCA-Net is capable of effectively disentangling domain-invariant features from domain-specific ones, extracting the domain-invariant features in a more structured and discriminative way, which is crucial for tasks requiring cross-domain generalization.

To demonstrate the effectiveness of the proposed FSCA-Net, we analyze the visual results of crowd counting and density prediction in the figure \ref{fig:density}. In terms of counting accuracy, by comparing predicted counts (PRE) with ground truth (GT) values, FSCA-Net+BL shows significant improvement over the baseline BL(M). For example, when GT is 1111, BL(M) predicts 1743 (severe overestimation), while FSCA-Net+BL predicts 1254, much closer to the truth; for GT = 297, BL(M) predicts 466 (substantial overestimation) versus FSCA-Net+BL’s 308 (nearly matching GT); and for GT = 1210, BL(M) predicts 1633 (severe overestimation) while FSCA-Net+BL predicts 1324, greatly reducing the error.

Regarding density prediction precision, BL(M)’s density heatmaps (Column C) exhibit inaccuracies, such as artificial high-density regions or failure to capture true dense areas. In contrast, FSCA-Net+BL’s heatmaps (Column D) closely align with the ground truth (Column B), accurately capturing the spatial distribution and intensity of dense regions (highlighted by black boxes). In summary, FSCA-Net not only enhances counting accuracy by reducing prediction errors but also improves density map precision by better capturing crowd spatial distribution, validating its effectiveness in crowd counting tasks.


\section{Conclusion}
In this work, we addressed the critical challenge of cross-domain crowd counting, where conventional CNN- and ViT-based models suffer from negative transfer due to the entanglement of domain-invariant and domain-specific features. To overcome this limitation, we proposed the Feature Separation and Cross-Attention Network (FSCA-Net), which explicitly disentangles feature spaces, enhances inter-domain interactions via cross-attention, and leverages mutual information constraints to suppress irrelevant variations. Comprehensive experiments on multiple benchmark datasets demonstrated that FSCA-Net significantly improves generalization while mitigating negative transfer, achieving state-of-the-art performance under cross-domain scenarios. Future work will extend this framework to real-time deployment and explore its applicability to other dense scene understanding tasks, such as vehicle counting and crowd behavior prediction.

\ifCLASSOPTIONcaptionsoff
  \newpage
\fi



\bibliographystyle{IEEEtran}
\bibliography{egbib}
\end{document}